\documentclass{article}

\usepackage{PRIMEarxiv}

\usepackage[utf8]{inputenc} 
\usepackage[T1]{fontenc}    
\usepackage{hyperref}       
\usepackage{url}            
\usepackage{booktabs}       
\usepackage{amsfonts}       
\usepackage{nicefrac}       
\usepackage{microtype}      
\usepackage{lipsum}
\usepackage{fancyhdr}       
\usepackage{graphicx}       
\usepackage{xcolor}
\graphicspath{{media/}}     
\usepackage{tablefootnote}
\usepackage{amsmath}
\usepackage{amssymb}
\usepackage{amsthm}
\title{Super Tiny Language Models}

\author{
  Leon Guertler \\
  A*STAR and Nanyang Technological University \\
  \texttt{leon002@e.ntu.edu.sg} \\
   \And
  Dylan Hillier \\
  A*STAR and Singapore Management University \\
  \texttt{das.hillier.2023@phdcs.smu.edu.sg} \\
   \AND
  {Palaash Agrawal, \ \ Chen Ruirui, \ \ Bobby Cheng, \ \ Cheston Tan} \\
  Centre for Frontier AI Research (CFAR), Institute of High Performance Computing (IHPC), A*STAR \\
  \texttt{\{agrawal\_palaash@cfar, chen\_ruirui@ihpc, bobby\_cheng@i2r, cheston\_tan@cfar\}.a-star.edu.sg} \\
}

\begin{document}
\maketitle

\begin{abstract}
The rapid advancement of large language models (LLMs) has led to significant improvements in natural language processing but also poses challenges due to their high computational and energy demands. This paper introduces a series of research efforts focused on Super Tiny Language Models (STLMs), which aim to deliver high performance with significantly reduced parameter counts. We explore innovative techniques such as byte-level tokenization with a pooling mechanism, weight tying, and efficient training strategies. These methods aim to significantly reduce reduce the parameter count compared to traditional models -- in future works, we aim to build on these in a way that maintains and improves upon the performance of base transformer models. This series of papers will explore into various subproblems, including tokenizer-free models, self-play based training, and alternative training objectives. We will target models with 10M, 50M, and 100M parameters. Our ultimate goal is to make high-performance language models more accessible and practical for a wide range of applications.
\end{abstract}


\section{Introduction}
\label{sec:introduction}
The recent rise in popularity of large language models (LLMs) has mostly been fueled by the invention of the attention based autoregressive transformer \cite{vaswani2017attention, radford2019language}. These models are trained to predict the next token (a subword unit) on very large corpora of data. There exists an extensive body of research on how scaling these architectures to more parameters and more data, predictably, improves model performance \cite{kaplan2020scaling, hestness2017deep, hoffmann2022training}. Thus, in practice, this is the recipe for performance improvement most companies (notably OpenAI, Google DeepMind, Anthropic) follow. However, with ever larger models, a number of problems arise. Firstly, the state-of-the-art models have been scaled to an extent where it is impossible for academic researchers to train competitive models, making safety and capability research harder. Secondly, training models of this magnitude requires so much compute and energy that industry players plan to build nuclear power plants, solely for the purpose of training models \cite{mark2024podcast, patterson2021carbon}. Lastly, when models are large, both inference time and applications that require running models on edge devices suffer.

Small language models on the other hand, on the order of 1 billion parameters, have been trained with increasingly impressive performance. This includes models like TinyLlama (1.1B)~\cite{zhang2024tinyllama}, Phi-3-mini (3.3B)~\cite{abdin2024phi}, and MobiLlama (0.5B)~\cite{thawakar2024mobillama}. While these models are able to reasonably compete with large foundation models, they still require thousands of GPU hours to train, which puts them out of the reach of many researchers, and thus prohibits fast experimentation.

Rather than focusing on recreating foundation models at a smaller scale, in this series of papers, we aim to use the small model size as a test bed for an open exploration of effective methods for improving parameter and sample efficiency. Specifically, we will focus on methods related to tokenizer-free models, weight tying, self-play based training, alternative training objectives, and data sampling techniques, 

\section{Related Works}
We provide basic information about transformers, especially as decomposed in our repository in Appendix~\ref{sec:llm_related}
In Section~\ref{sec:param_red_related_work} we cover some of the works on parameter reduction that inform our research

\subsection{Parameter Reduction}
\label{sec:param_red_related_work}
In the quest to make language models more efficient and accessible, various parameter reduction techniques have been developed. These techniques aim to reduce the number of parameters in the model without significantly compromising performance. Below, we discuss some of the most popular methods, including weight tying, pruning, quantization, and knowledge distillation. We note that many of these techniques are primarily used in a post-hoc fashion and may not be applicable during model training.

\paragraph{Weight Tying}
Weight tying is a technique where certain weights in the model are shared between different components. This approach not only reduces the total number of parameters but also ensures that certain parts of the model are better aligned. There are different types of weight tying used in various models:
\begin{itemize}
    \item \textbf{Embedding \& Head} In GPT-2 and other similar models, the embedding matrix is tied to the weights of the output layer, ensuring that the output probabilites are directly related to the input embeddings 
    \item \textbf{FFN sharing} MobiLlama \cite{thawakar2024mobillama} shares weights specifically between the feed-forward network (FFN) layers. By doing so, it achieves parameter efficiency without compromising on the model's ability to learn and generalize.
    \item \textbf{FFN+Attn sharing} ALBERT \cite{lan2019albert} employs weight tying extensively by sharing parameters across all layers of the transformer. It ties the weights of both the feed-forward network (FFN) and the attention layers, which significantly reduces the model size while maintaining performance.\cite{press2017using}.

\end{itemize}

\paragraph{Pruning} Pruning involves removing weights that contribute least to the model's performance. This can be done during or after training. Pruning results in a sparser model with fewer parameters and reduced computational requirements. This is inspired by the \textit{lottery ticket hypothesis} which states that there exists a smaller subnetwork (a ``winning ticket'') that, when trained in isolation, can achieve performance comparable to the original model \cite{frankle2018lottery}. Pruning methods inspired by this hypothesis identify and retain only the most critical parameters.

\paragraph{Quantization} Quantization reduces the precision of the model's weights and activations from 32-bit floating-point numbers to lower-bit representations such as 8-bit integers. This technique significantly reduces model size (if not parameter count) and often speeds up training/inference with minimal impact on performance.\cite{jacob2018quantization}

\paragraph{Low-Rank Factorization} This technique decomposes large weight matrices into products of smaller matrices, which reduces the number of parameters and the computational cost.
This has been used in e.g. Ma et al.\cite{ma2019tensorized} for compressing a pretrained BERT model

\subsection{Data Quality and Training Efficiency} In place of directly reducing the number of parameters, other approaches focus on improving the quality of the training signal, and thereby enabling the use of fewer parameters. Below, we cover data selection and knowledge distillation.

\paragraph{Data Selection}
A key argument of the Phi series of language models\cite{abdin2024phi,javaheripi2023phi, gunasekar2024textbooks} is that by improving the quality of data, the performance of small language models can be increased, far in excess of the predictions of e.g. scaling laws. In contrast to training on crawls of the internet, these models are trained on textbooks and heavily filtered web data, and are able to match performance with larger models trained on larger training sets. Some data augmentation and filtering techniques go further to use pretrained large language models as part of the process. For example Zhang et al.~\cite{zhang2024autonomous} utilise a pretrained language model to verify the quality of training data as a mathematic source during finetuning.

\paragraph{Knowledge Distillation} Knowledge distillation transfers the knowledge from a larger ``teacher'' model to a smaller ``student'' model. In particular the idea is that rather than the model being trained on a single hard label (i.e. a particular token), the student model instead has the entire probability distribution of the teacher model to guide its learning. For example, DistilBERT~\cite{sanh2019distilbert} is a smaller version of BERT created using knowledge distillation which retains $97\%$ of BERT's language understanding capabilities while being $60\%$ faster and $40\%$ smaller .

\section{Statement of Goals / Measurements of Success}
Whilst the end-goal is to create highly performant super tiny language models, we measure our progress towards that goal along multiple dimensions, including:

\paragraph{Model Size} We propose no hard parameter count cut-off, but aim to build a family of models with 10M, 50M and 100M parameters. Most experiments will focus on models with close to 50M parameters, and once we have narrowed down an architecture, we will scale it up and down to 100M and 10M parameters respectively.

\paragraph{Training Time} As mentioned in the introduction, one of the key motivators for this series of papers is to make research more accessible. For this to work, training time on consumer architecture needs to be short enough to allow for experimentation. Thus, for the 50M model, the overall training time should be less than 48 GPU Hours, which enables researchers to reasonably experiment even with a single GPU.

\paragraph{Model Performance} To check whether our STLMs achieve competitive results despite their reduced size, we will evaluate their performance across a range of benchmarks and metrics commonly used for evaluating much larger models. As described in Section~\ref{sec: eval} this will include MMLU~\cite{hendrycks2021measuring}, HellaSwag~\cite{zellers2019hellaswag}, and ARC~\cite{clark2018think}. In the long term, in order to compare the usability of these models for downstream tasks we aim to produce small models that are reasonably competitive with 3-7B parameter models on benchmarks like GSM8K~\cite{cobbe2021training} and LMSYS Chatbot Arena~\cite{chiang2024chatbot} after instruction tuning.

\section{Proposed Approach}
\label{sec:proposed_structure}
\subsection{Technical}
Our technical contribution is a research repository in which we aim to enable other researchers to easily run experiments on small models by surfacing clean, accessible interfaces, with interpretable and understandable model code. This is implemented in PyTorch~\cite{paszke2017automatic}, with Hydra for config management, and logging in Weights and Biases. The repository is initially based off of Karpathy's minGPT repository~\cite{karpathy2022mingpt}, but we also implement the so-called ``modern transformer''~\cite{gu2023mamba} used in e.g. Llama 2~\cite{touvron2023llama}. As such, the core components exposed in our library consist of:
\begin{itemize}
    \item GPT2 / BPE Tokenizers
    \item SwiGLU / Standard FFN layers
    \item Attention supporting RoPE embeddings, Causal Masks, Grouped Query Attention
\end{itemize}
\subsubsection{Training Data}
\label{sec: train_data}
For the set of baselines presented here we use OpenWebText~\cite{Gokaslan2019OpenWeb} as our training data due to its high diversity and coverage. This constitutes around 6.5 billion words of data (on the order of 60 times the number seen by a 12 year-old~\cite{warstadt2023findings}). We run these baselines for 50,000 steps. 
In our initial training runs, we used SimpleEnglishWiki as our training data and found that this began to overfit after 5,000 epochs. This is somewhat unsurprising given that during this training process the model sees on the order of $3$ billion words -- in other words, the model sees each word in the training dataset 100 times. Indeed, per the Kaplan Scaling laws~\cite{kaplan2020scaling} a model of this size requires roughly 1 billion words to avoid significant fitting. This was the motivation for our move to OpenWebText, however we note that a number of works have shown the significant impact of data quality, motivating further research on the topic.
As such we intend to investigate the effects of different training data sets, including:
\begin{itemize}
    \item Simple English Wikipedia 
    \item English Wikipedia
    \item BabyLM~\cite{warstadt2023findings}
    \item A mix containing simple english wiki, a python code contest datase\footnote{\url{https://huggingface.co/datasets/jtatman/python-code-dataset-500k}}, the OpenHermes chat dataset\cite{OpenHermes2.5}, and the TinyStories dataset\cite{eldan2023tinystories}.
\end{itemize}

\subsubsection{Evaluation}
\label{sec: eval}
The usual measure of language modelling performance for a model with probability distribution $P$, sequence $\mathbf{x}$ is given by the perplexity:
\[\log_2{\left(\text{Perplexity}\right)} = - \frac{1}{N} \sum_{i=1}^{N} \log_2(P(x_i|\mathbf{x}_{:i}))
\] One intuitive way of thinking of the perplexity is to measure how many guesses are required on average per prediction step, the issue being that for different tokenizers, there will be different sized prediction steps. In this project, since we aim to enable experiments across different tokenizers, we need to equalize this measure by using the original string length, rather than the length of the input string. Perplexity is evaluated on a held out test set from the training data.

In addition, we initially use the following as question answering datasets for evaluating performance:
\begin{itemize}
    \item BLiMP~\cite{warstadt2020blimp} -- Instead of measuring reasoning, the Benchmark of Linguistic Minimal Pairs tests the grammatical understanding of models across a range of 67 grammatical phenomena in English.
    \item HellaSwag~\cite{zellers2019hellaswag} -- Short for Harder Endings, Longer Contexts and Low-shot Activities for Situations with Adversarial Generations, HellaSwag measures a language model's natural language inferencing (NLI) capabilities. It does so by feeding the language model with incomplete events, and expects it to complete the story with the most sensible answer.
    \item ARC~\cite{clark2018think} -- The AI2 Reasoning Challenge (ARC) is aimed at evaluating a language model's knowledge and reasoning skills. Unlike other question and answering benchmarks like the Stanford Question and Answer Dataset (SQuAD), which can be solved by explicitly remembering answers, ARC is meant to be a more challenging and extensive benchmark by forcing the language model to respond to questions using its learnt knowledge and reasoning abilities.
    \item WinoGrande~\cite{sakaguchi2021winogrande} -- This benchmark evaluates a language model's ability to generate common sense reasoning by solving 'pronoun resolution problems'. Concretely, it feeds the language model with pairs of input sentences which differ only in the interpretation of particular pronoun. High performance indicates the models ability to use semantics to resolve ambiguities.
    \item MMLU~\cite{hendrycks2021measuring} -- The Massive Multitask Language Understanding (MMLU) benchmark assesses a language model's understanding of a wide-range of subject topics that span from general knowledge to problem solving capabilities. By covering a wide range of topics like technology, science and mathematics, it tells us how well the language model can understand a given text, and how well informed it is from its training procedures.

\end{itemize}
These are all multiple choice QA datasets. We use the path probabilities over the different options to measure which option is ``preferred'' by the model, which is intended to mitigate the lack of instruction following/question answering in the training data and thus better measure performance disparities of these models.
\begin{figure}[ht]
    \centering
    \includegraphics[width=0.9\linewidth]{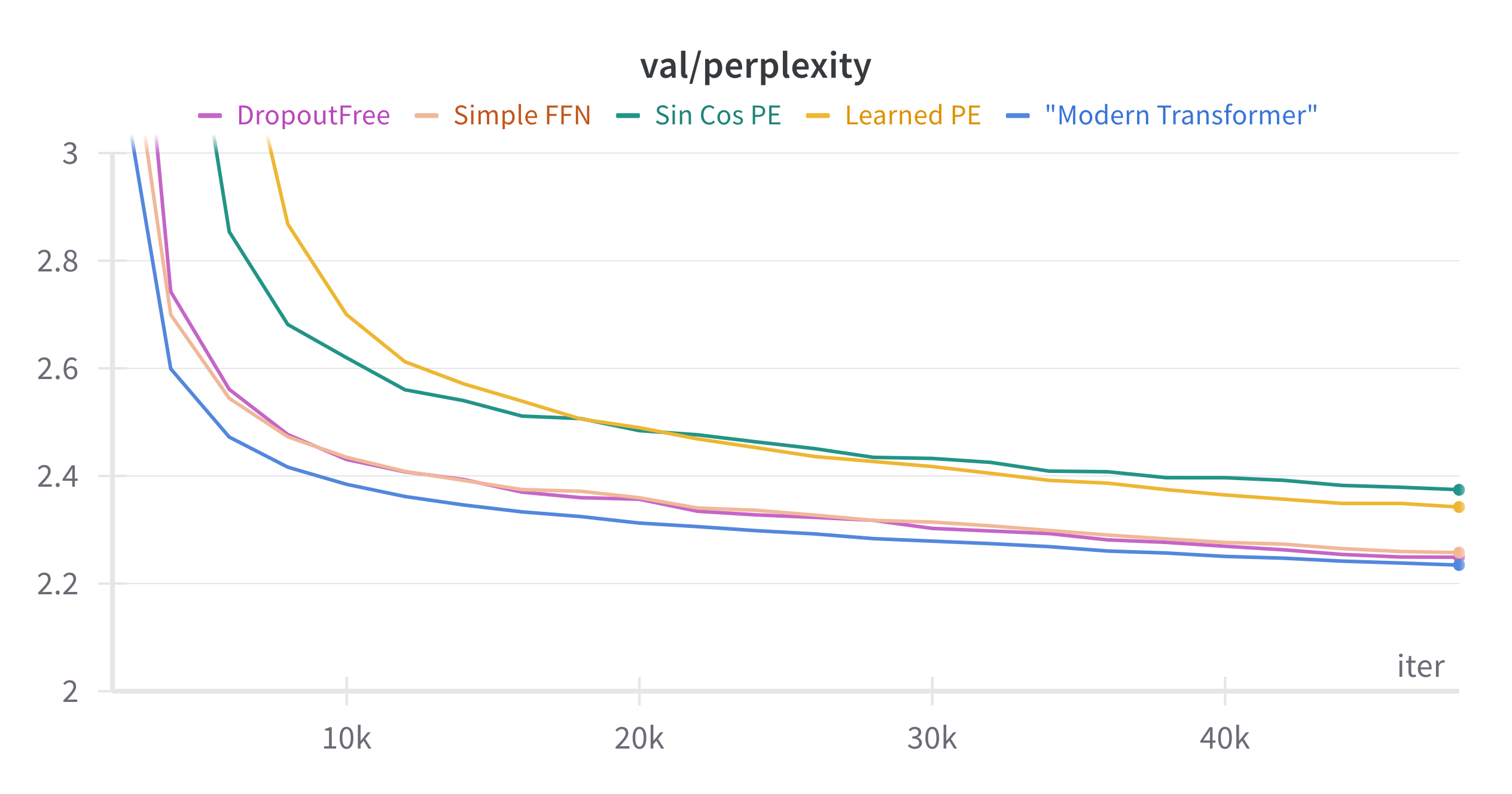}
    \caption{Perplexity during training for the baseline models}
    \label{fig:perplexity curves}
\end{figure}
\begin{table}[ht]
\centering
\begin{tabular}{ll}
\toprule
\multicolumn{2}{c}{Model} \\
\midrule
number of layers & 8 \\
ffn type & SwiGLU \\
ffn dimension & 1536 \\
number of attention heads & 16 \\
group size & 4 \\
tokenizer type & gpt2 \\
hidden dim & 512 \\
max context window & 512 \\
vocab\_size & 50257 \\
\midrule
\multicolumn{2}{c}{Training} \\
\midrule
batch\_size & 24 \\
gradient accumulation\_steps & 20 \\
total iterations & 25000 \\
warmup iterations & 5000 \\
dropout & 0.1 \\
\midrule
\multicolumn{2}{c}{Optimizer} \\
\midrule
lr & 0.0006 \\
weight\_decay & 0.1 \\
\midrule
\multicolumn{2}{c}{Miscellaneous} \\
\midrule
GPU & 2xL40 \\
Total GPU Hours & 30 \\
\bottomrule
\end{tabular}
\caption{
The parameters used for our baseline
}
\label{table: Baseline Config}
\end{table}
\begin{table}[]
    \centering
    \begin{tabular}{lcccccc}
\toprule
Model & Num. Params & BLiMP & HellaSwag & ARC\_easy & WinoGrande & MMLU \\ \midrule
Phi-3                                  & 3B   & 83\% & 56\% & 80\% & 73\% & 32\% \\
TinyLlama~\cite{zhang2024tinyllama}    & 1B   & 84\% & 43\% & 60\% & 59\% & 26\% \\
MobiLlama~\cite{thawakar2024mobillama} & 500M & 80\% & 39\% & 51\% & 55\% & 25\% \\ 
Qwen2~\cite{qwen}                      & 500M & 80\% & 38\% & 52\% & 55\% & 28\% \\
\midrule
Baseline                               & 50M  & 78\% & 30\% & 39\% & 50\% & 24\% \\
\midrule
SinCos PE                              & 47M  & 77\% & 29\% & 35\% & 50\% & 24\% \\
Learned PE                             & 48M  & 77\% & 29\% & 37\% & 50\% & 23\% \\
Simple FFN                             & 50M  & 79\% & 30\% & 39\% & 51\% & 24\% \\
DropoutFree                            & 47M  & 77\% & 30\% & 39\% & 51\% & 24\% \\
\midrule
Metric & -- &Accuracy & Accuracy & Accuracy & Accuracy & Accuracy\\
Num. Choices                           & --   & 2    & 4    & 4    & 2    & 4    \\
Chance Perf.                           &  --  & 50\% & 25\% & 25\% & 50\% & 25\% \\
\bottomrule
\end{tabular}
    \caption{Initial results on chosen benchmarks. All results are obtained zero-shot with custom splits of the datasets.}
    \label{tab: init results}
\end{table}

\subsection{Benchmarking}
We run benchmarks with a number of 50M parameter models. The baseline values are given in Table~\ref{table: Baseline Config}. In particular `Sin Cos PE' uses the positional embeddings from the original transformer paper~\cite{vaswani2017attention}; `Learned PE' just uses learnt positional embeddings; and `Simple FFN' uses a 2 layer feed forward network with GeLU~\cite{hendrycks2016gaussian} activations. This is in contrast to the `Modern Transformer' baseline which has RoPE~\cite{su2024roformer} positional embeddings, and a SwiGLU~\cite{shazeer2020glu} FFN structure. Finally, we run a `DropoutFree' version with 0 dropout, observing that this has lower perplexity, and comparable eval scores despite being smaller. We note that for all of these baseline models, the embeddings take up 25M parameters, i.e. over half the total parameter allocation.

In Figure~\ref{fig:perplexity curves} and Table~\ref{tab: init results} we show the initial results on the benchmarks described in Section~\ref{sec: eval}. Preliminarily, we reinforce the value of using RoPE embeddings, but note that while the validation perplexity on the `Simple FFN' model is higher, it nevertheless performs incrementally better across the board on our benchmarks. This requires further experimentation to validate however as the feedforward dimension used here (2304) was much higher than in the baseline (1536). The `width' of a transformer model~\cite{petty2024impact} can have a significant impact on the performance of the downstream model, and as such our rather cavalier altering of the FFN dimension makes it difficult to directly draw conclusions.

More generally, we can see that these baselines are around chance at MMLU and WinoGrande (although the <1B models are not far above chance). On BLiMP, the baselines are fairly close to the 500M parameter models, while for ARC-easy and HellaSwag the performance gap is reasonable but presents a reasonable target for this series of works.

\subsection{Research}
In order to increase the validity of our research we plan on committing to releasing reports regardless of the outcome, and release negative results alongside positive ones.

\section{Proposed Research Projects}
In line with the research goals described above, we outline a number of research projects that we aim to explore in the coming months.

\subsection{Weight Tying}\label{sec:weighttying}
As introduced in section~\ref{sec:weighttying}, weight tying is a commonly used method of parameter reduction. Recent methods for creating tiny language models including MobiLlama\cite{thawakar2024mobillama} have gone further to share the weights between the feed forward layers. Since the majority of transformer parameters are stored in these layers this makes sense as a target for parameter reduction, additionally recent research has shown that many of these layers are redundant\cite{men2024shortgpt}. We currently propose to explore different styles of feedforward weight sharing, including using LoRA adaptors\cite{hu2021lora} to allow for parameter efficient adaptation of the computation performed by these layers.

\subsection{Byte-level / tokenizer free}
One of the most parameter expensive components of the modern day transformer architecture are the embedding layer and the next token head. The reason for the high parameter count in these two layers is the number of unique tokens produced by the tokenizer. The size of the this embedding layer is \text{vocabulary size} * \text{embedding dimension}, where the vocabulary size is typically on the order of 50000 tokens (e.g. 50,257 for GPT2~\cite{brown2020language}). This is then replicated in the LM head, which in some cases may not share the weights of the embedding layer e.g. in Press et al.~\cite{press2017using}. As such for a GPT-2 sized model, the embedding and lm head layers can account for $45.20\%$ to $62.25\%$ of total parameter count depending on whether weight tying is used.

One solution to this is finding some sort of architecture that requires a significantly smaller vocabulary size (in the extreme this could be $256$ for byte level models), however, as the vocabulary size shrinks, it becomes harder for the model to fit the data, and efficiency degrades (as the sequence lengths (in terms of token count) increase for the same input string). We propose to explore a byte-level tokenizer with a pooling mechanism. Specifically, the proposed method works as follows:
\begin{itemize}
    \item \textbf{Byte-level Embedding} The input string is first embedded using a byte-level embedder. Each character/byte in the input string is converted into a byte representation, resulting in a sequence of byte embeddings. Since the vocabulary size is only $256$, a very small embedding dimension can be used ($64$ for example), this step only requires a negligible number of trainable parameters compared to the total model.
    \item \textbf{Tokenization and Chunking} The byte-embedded sequence is then split into chunks using the bounding boxes of a standard Byte Pair Encoding (BPE) tokenizer. Importantly, this step requires no additional parameters as the tokenizer is only used to find the bounding boxes of tokens and not to embed the input.
    \item \textbf{Pooling Mechanism} Based on the bounding boxes obtained from the BPE tokenizer, the byte embeddings are chunked into tokens. Each chunk is then individually passed through a smaller two-layer transformer. The output of this transformer is pooled into a single token representation. At this point the dimensions of the hidden state are the exact same it would be if a standard BPE tokenizer were used, however, this method requires only a fraction ($~10\%$) or the parameters.
    \item \textbf{Core Model Processing} The pooled token representations are then passed through the core model. The dimension of the data going through the core model is the same as it would be if just a BPE tokenizer was used.
    \item \textbf{Decoding and Final Prediction} Instead of using a language model head, each BPE-level token is decoded back into bytes. The final loss is calculated as the next BPE token prediction but for each byte in the next BPE token. This is done via a similar mechanism as the byte-level encoder, and thus, also requires substantially fewer parameters.
\end{itemize}

Overall, compared to using a BPE tokenizer, the proposed Byte-level model uses between $90\%$ and $95\%$ fewer parameters, depending on whether weight-tying is used or not.

\subsection{Early Exit and Conditional Computation}
The success of techniques like Mixture of Experts e.g.~\cite{jiang2024mixtral} is thought to lie in the fact that different tokens may require different computation pathways. More generally we can expect that different tokens require different amounts of computation. As argued in the previous section, deeper layers of transformers may be redundant for a majority of inputs -- we aim to exploit this by exploring methods that skip this computation for easy to predict tokens. In particular we plan to explore mixture of depths~\cite{raposo2024mixture} and the recently proposed layerskip~\cite{elhoushi2024layer} methods.

\subsection{Next thought prediction}
Again building on the idea of conditional computation, we are interested in exploring methods that perform computation over sequences of thoughts rather than sequences of test - more generally decoupling the language modelling capabilities from the reasoning capabilities. Language models are of course primarily designed to model language, unfortunately this means that while they have impressive emergent downstream capabilities~\cite{brown2020language} and (arguably) various interesting properties~\cite{kosinski2023theory} these firstly may not emerge at smaller model scales and furthermore may occur in spite of rather than because of the language modelling training paradigm. In particular per the arguments of LeCun et al.~\cite{lecun2022path}, it may be more efficient to perform predictions in a latent space that is not directly tied to the output sequence space. Indeed this is similar to the way that Chain-of-thought~\cite{wei2022chain} augments a model's downstream capabilities by leveraging the models knowledge of reasoning rather than directly answering questions. Thus for example, Goyal et al.~\cite{goyal2023think} utilise pause tokens that do not correspond to tokens in the source text, and enable the model to perform additional computation before outputting an answer.

\subsection{Dropout and Learning Rate Scheduling}
We would like to explore the effect of using dropout. While most large models are not trained with dropout due to using sufficiently large datasets as to avoid over-fitting, Liu et al~\cite{liu2023dropout} proposes that scheduling dropout in the early phases of training can help reduce underfitting (which may be relevant when training on small amounts of high quality data) and if scheduled for the late phases is more effective at combating overfitting. Additionally most approaches, such as Llama~\cite{touvron2023llama} use Cosine Learning Rate Schedulers. Given the relatively fast training times it would be useful to verify the efficacy of this scheduler.

\subsection{Curriculums, Data mixes, and Multimodality}
As discussed in section~\ref{sec: train_data}, the quality of the training data can greatly impact the performance of the trained model. Additionally having training data that coverages a sufficient portion of use cases (i.e. not just factual wikipedia articles) is necessary for drawing reasonable inferences from the downstream performances of tiny llms. Measuring the quality and coverage of training data is difficult however, especially for foundation models where the downstream tasks may be unclear. One potential option is to use high quality datasets like the British National Corpus~\cite{leech1992100} with a sufficiently broad set of data sources, or to use a refined/ heavily filtered version of webcrawls in the manner of Penedo et al.~\cite{penedo2024refinedweb}


\section{Conclusion}
\label{sec:conclusion}
In this introductory paper, we have outlined the vision and approach for developing Super Tiny Language Models (STLMs) aimed at achieving high performance with significantly reduced parameter counts. The growing computational and energy demands of large language models underscore the necessity of this research. Our proposed methods, including byte-level tokenization with pooling, weight tying, and efficient training strategies, have the potential to reduce the parameter count by 90\% to 95\% compared to traditional models while maintaining competitive performance.

We have detailed various techniques for parameter reduction, such as weight tying, pruning, quantization, and knowledge distillation, and discussed their relevance to our goals. Our technical approach involves creating a research repository that facilitates experimentation with small models, making cutting-edge NLP research more accessible.

The performance of our STLMs will be rigorously evaluated using standard benchmarks, aiming to match or exceed the capabilities of much larger models. This series of papers will delve into specific subproblems, including tokenizer-free models, self-play based training, and alternative training objectives, targeting models with 10M, 50M, and 100M parameters.

Ultimately, our goal is to democratize access to high-performance language models, enabling more researchers and practitioners to contribute to and benefit from advancements in NLP. We believe that the development of STLMs will pave the way for more sustainable and efficient AI, broadening the scope of applications and fostering innovation across various domains.

We invite the research community to engage with our work, explore the proposed techniques, and contribute to the ongoing effort to make language models more efficient and widely usable.

\bibliographystyle{unsrt}  
\bibliography{references}  

\begin{thebibliography}{10}

\bibitem{vaswani2017attention}
Ashish Vaswani, Noam Shazeer, Niki Parmar, Jakob Uszkoreit, Llion Jones, Aidan~N Gomez, Lukasz Kaiser, and Illia Polosukhin.
\newblock Attention is all you need.
\newblock {\em Advances in neural information processing systems}, 30, 2017.

\bibitem{radford2019language}
Alec Radford, Jeffrey Wu, Rewon Child, David Luan, Dario Amodei, and Ilya Sutskever.
\newblock Language models are unsupervised multitask learners.
\newblock 2019.

\bibitem{kaplan2020scaling}
Jared Kaplan, Sam McCandlish, Tom Henighan, Tom Brown, Benjamin Chess, Rewon Child, Scott Gray, Alec Radford, Jeffrey Wu, and Dario Amodei.
\newblock Scaling laws for neural language models.
\newblock {\em arXiv preprint arXiv:2001.08361}, 2020.

\bibitem{hestness2017deep}
Joel Hestness, Sharan Narang, Niki Ardalani, Greg Diamos, Heewoo Jun, Hessam Kianinejad, Mostofa Patwary, Mohammad~Shoeybi Ali, Erich Yong, and Stephen Zhu.
\newblock Deep learning scaling is predictable, empirically.
\newblock {\em arXiv preprint arXiv:1712.00409}, 2017.

\bibitem{hoffmann2022training}
Jordan Hoffmann, Sebastian Borgeaud, Arthur Mensch, Elena Buchatskaya, Trevor Cai, Eliza Rutherford, Diego de~Las Casas, Aurelia Guy, Sarah Henderson, Katie Millican, et~al.
\newblock Training compute-optimal large language models.
\newblock {\em arXiv preprint arXiv:2203.15556}, 2022.

\bibitem{mark2024podcast}
Mark~Zuckerberg Dwarkesh~Patel.
\newblock Mark zuckerbart - llama 3, open sourcing 10b models, \& caesar augustus.
\newblock Podcast, 2024.
\newblock Accessed: 2024-05-19, URL: https://open.spotify.com/episode/6Lbsk4HtQZfkJ4dZjh7E7k?si=c31a37a19f994175.

\bibitem{patterson2021carbon}
David Patterson, Joseph Gonzalez, Quoc Le, Chen Liang, Lluis-Miquel Munguia, Daniel Rothchild, David So, Morgan Texier, and Jeff Dean.
\newblock Carbon emissions and large neural network training.
\newblock {\em arXiv preprint arXiv:2104.10350}, 2021.

\bibitem{zhang2024tinyllama}
Peiyuan Zhang, Guangtao Zeng, Tianduo Wang, and Wei Lu.
\newblock Tinyllama: An open-source small language model.
\newblock {\em arXiv preprint arXiv:2401.02385}, 2024.

\bibitem{abdin2024phi}
Marah Abdin, Sam~Ade Jacobs, Ammar~Ahmad Awan, Jyoti Aneja, Ahmed Awadallah, Hany Awadalla, Nguyen Bach, Amit Bahree, Arash Bakhtiari, Harkirat Behl, et~al.
\newblock Phi-3 technical report: A highly capable language model locally on your phone.
\newblock {\em arXiv preprint arXiv:2404.14219}, 2024.

\bibitem{thawakar2024mobillama}
Omkar Thawakar, Ashmal Vayani, Salman Khan, Hisham Cholakal, Rao~M Anwer, Michael Felsberg, Tim Baldwin, Eric~P Xing, and Fahad~Shahbaz Khan.
\newblock Mobillama: Towards accurate and lightweight fully transparent gpt.
\newblock {\em arXiv preprint arXiv:2402.16840}, 2024.

\bibitem{lan2019albert}
Zhenzhong Lan, Mingda Chen, Sebastian Goodman, Kevin Gimpel, Piyush Sharma, and Radu Soricut.
\newblock Albert: A lite bert for self-supervised learning of language representations.
\newblock {\em arXiv preprint arXiv:1909.11942}, 2019.

\bibitem{press2017using}
Ofir Press and Lior Wolf.
\newblock Using the output embedding to improve language models, 2017.

\bibitem{frankle2018lottery}
Jonathan Frankle and Michael Carbin.
\newblock The lottery ticket hypothesis: Finding sparse, trainable neural networks.
\newblock {\em arXiv preprint arXiv:1803.03635}, 2018.

\bibitem{jacob2018quantization}
Benoit Jacob, Skirmantas Kligys, Bo~Chen, Menglong Zhu, Matthew Tang, Andrew Howard, Hartwig Adam, and Dmitry Kalenichenko.
\newblock Quantization and training of neural networks for efficient integer-arithmetic-only inference.
\newblock {\em Proceedings of the IEEE Conference on Computer Vision and Pattern Recognition}, pages 2704--2713, 2018.

\bibitem{ma2019tensorized}
X~Ma, J~Zhang, R~Wang, Q~Xu, and D~Lin.
\newblock Tensorized embedding layers for efficient model compression.
\newblock {\em Advances in Neural Information Processing Systems}, 32, 2019.

\bibitem{javaheripi2023phi}
Mojan Javaheripi, S{\'e}bastien Bubeck, Marah Abdin, Jyoti Aneja, Sebastien Bubeck, Caio C{\'e}sar~Teodoro Mendes, Weizhu Chen, Allie Del~Giorno, Ronen Eldan, Sivakanth Gopi, et~al.
\newblock Phi-2: The surprising power of small language models.
\newblock {\em Microsoft Research Blog}, 2023.

\bibitem{gunasekar2024textbooks}
Suriya Gunasekar, Yi~Zhang, Jyoti Aneja, Caio Cesar~Teodoro Mendes, Allie~Del Giorno, Sivakanth Gopi, Mojan Javaheripi, Piero~Conti Kauffmann, Gustavo~Henrique de~Rosa, Olli Saarikivi, Adil Salim, Shital Shah, Harkirat Behl, Xin Wang, Sebastien Bubeck, Ronen Eldan, Adam~Tauman Kalai, Yin~Tat Lee, and Yuanzhi Li.
\newblock Textbooks are all you need, 2024.

\bibitem{zhang2024autonomous}
Yifan Zhang, Yifan Luo, Yang Yuan, and Andrew~C Yao.
\newblock Autonomous data selection with language models for mathematical texts.
\newblock In {\em ICLR 2024 Workshop on Navigating and Addressing Data Problems for Foundation Models}, 2024.

\bibitem{sanh2019distilbert}
Victor Sanh, Lysandre Debut, Julien Chaumond, and Thomas Wolf.
\newblock Distilbert, a distilled version of bert: smaller, faster, cheaper and lighter.
\newblock {\em arXiv preprint arXiv:1910.01108}, 2019.

\bibitem{hendrycks2021measuring}
Dan Hendrycks, Collin Burns, Steven Basart, Andy Zou, Mantas Mazeika, Dawn Song, and Jacob Steinhardt.
\newblock Measuring massive multitask language understanding.
\newblock In {\em International Conference on Learning Representations}, 2021.

\bibitem{zellers2019hellaswag}
Rowan Zellers, Ari Holtzman, Yonatan Bisk, Ali Farhadi, and Yejin Choi.
\newblock Hellaswag: Can a machine really finish your sentence?
\newblock {\em arXiv preprint arXiv:1905.07830}, 2019.

\bibitem{clark2018think}
Peter Clark, Isaac Cowhey, Oren Etzioni, Tushar Khot, Ashish Sabharwal, Carissa Schoenick, and Oyvind Tafjord.
\newblock Think you have solved question answering? try arc, the ai2 reasoning challenge.
\newblock {\em arXiv preprint arXiv:1803.05457}, 2018.

\bibitem{cobbe2021training}
Karl Cobbe, Vineet Kosaraju, Mohammad Bavarian, Alhussein Fawzi, Felix Chen, Mateusz Malinowski, Sanja Fidler, Oleg Klimov, Tim Vogels, Jacob Hilton, Jessica Han, and John Schulman.
\newblock Training verifiers to solve math word problems.
\newblock {\em arXiv preprint arXiv:2106.07311}, 2021.

\bibitem{chiang2024chatbot}
Wei-Lin Chiang, Lianmin Zheng, Ying Sheng, Anastasios~Nikolas Angelopoulos, Tianle Li, Dacheng Li, Hao Zhang, Banghua Zhu, Michael Jordan, Joseph~E. Gonzalez, and Ion Stoica.
\newblock Chatbot arena: An open platform for evaluating llms by human preference, 2024.

\bibitem{paszke2017automatic}
Adam Paszke, Sam Gross, Soumith Chintala, Gregory Chanan, Edward Yang, Zachary DeVito, Zeming Lin, Alban Desmaison, Luca Antiga, and Adam Lerer.
\newblock Automatic differentiation in pytorch.
\newblock In {\em NIPS-W}, 2017.

\bibitem{karpathy2022mingpt}
Andrej Karpathy.
\newblock mingpt.
\newblock \url{https://github.com/karpathy/minGPT}, 2022.
\newblock GitHub repository.

\bibitem{gu2023mamba}
Albert Gu and Tri Dao.
\newblock Mamba: Linear-time sequence modeling with selective state spaces.
\newblock {\em arXiv preprint arXiv:2312.00752}, 2023.

\bibitem{touvron2023llama}
Hugo Touvron, Louis Martin, Kevin Stone, Peter Albert, Amjad Almahairi, Yasmine Babaei, Nikolay Bashlykov, Soumya Batra, Prajjwal Bhargava, Shruti Bhosale, et~al.
\newblock Llama 2: Open foundation and fine-tuned chat models.
\newblock {\em arXiv preprint arXiv:2307.09288}, 2023.

\bibitem{Gokaslan2019OpenWeb}
Aaron Gokaslan and Vanya Cohen.
\newblock Openwebtext corpus.
\newblock \url{http://Skylion007.github.io/OpenWebTextCorpus}, 2019.

\bibitem{warstadt2023findings}
Alex Warstadt, Aaron Mueller, Leshem Choshen, Ethan Wilcox, Chengxu Zhuang, Juan Ciro, Rafael Mosquera, Bhargavi Paranjabe, Adina Williams, Tal Linzen, et~al.
\newblock Findings of the babylm challenge: Sample-efficient pretraining on developmentally plausible corpora.
\newblock In {\em Proceedings of the BabyLM Challenge at the 27th Conference on Computational Natural Language Learning}, 2023.

\bibitem{OpenHermes2.5}
Teknium.
\newblock Openhermes 2.5: An open dataset of synthetic data for generalist llm assistants, 2023.

\bibitem{eldan2023tinystories}
Ronen Eldan and Yuanzhi Li.
\newblock Tinystories: How small can language models be and still speak coherent english?
\newblock {\em arXiv preprint arXiv:2305.07759}, 2023.

\bibitem{warstadt2020blimp}
Alex Warstadt, Alicia Parrish, Haokun Liu, Anhad Mohananey, Wei Peng, Sheng-Fu Wang, and Samuel~R Bowman.
\newblock Blimp: The benchmark of linguistic minimal pairs for english.
\newblock {\em Transactions of the Association for Computational Linguistics}, 8:377--392, 2020.

\bibitem{sakaguchi2021winogrande}
Keisuke Sakaguchi, Ronan~Le Bras, Chandra Bhagavatula, and Yejin Choi.
\newblock Winogrande: An adversarial winograd schema challenge at scale.
\newblock {\em Communications of the ACM}, 64(9):99--106, 2021.

\bibitem{qwen}
Jinze Bai, Shuai Bai, Yunfei Chu, Zeyu Cui, Kai Dang, Xiaodong Deng, Yang Fan, Wenbin Ge, Yu~Han, Fei Huang, Binyuan Hui, Luo Ji, Mei Li, Junyang Lin, Runji Lin, Dayiheng Liu, Gao Liu, Chengqiang Lu, Keming Lu, Jianxin Ma, Rui Men, Xingzhang Ren, Xuancheng Ren, Chuanqi Tan, Sinan Tan, Jianhong Tu, Peng Wang, Shijie Wang, Wei Wang, Shengguang Wu, Benfeng Xu, Jin Xu, An~Yang, Hao Yang, Jian Yang, Shusheng Yang, Yang Yao, Bowen Yu, Hongyi Yuan, Zheng Yuan, Jianwei Zhang, Xingxuan Zhang, Yichang Zhang, Zhenru Zhang, Chang Zhou, Jingren Zhou, Xiaohuan Zhou, and Tianhang Zhu.
\newblock Qwen technical report.
\newblock {\em arXiv preprint arXiv:2309.16609}, 2023.

\bibitem{hendrycks2016gaussian}
Dan Hendrycks and Kevin Gimpel.
\newblock Gaussian error linear units (gelus).
\newblock {\em arXiv preprint arXiv:1606.08415}, 2016.

\bibitem{su2024roformer}
Jianlin Su, Murtadha Ahmed, Yu~Lu, Shengfeng Pan, Wen Bo, and Yunfeng Liu.
\newblock Roformer: Enhanced transformer with rotary position embedding.
\newblock {\em Neurocomputing}, 568:127063, 2024.

\bibitem{shazeer2020glu}
Noam Shazeer.
\newblock Glu variants improve transformer.
\newblock {\em arXiv preprint arXiv:2002.05202}, 2020.

\bibitem{petty2024impact}
Jackson Petty, Sjoerd Steenkiste, Ishita Dasgupta, Fei Sha, Dan Garrette, and Tal Linzen.
\newblock The impact of depth on compositional generalization in transformer language models.
\newblock In {\em Proceedings of the 2024 Conference of the North American Chapter of the Association for Computational Linguistics: Human Language Technologies (Volume 1: Long Papers)}, pages 7232--7245, 2024.

\bibitem{men2024shortgpt}
Xin Men, Mingyu Xu, Qingyu Zhang, Bingning Wang, Hongyu Lin, Yaojie Lu, Xianpei Han, and Weipeng Chen.
\newblock Shortgpt: Layers in large language models are more redundant than you expect.
\newblock {\em arXiv preprint arXiv:2403.03853}, 2024.

\bibitem{hu2021lora}
Edward~J Hu, Yelong Shen, Phillip Wallis, Zeyuan Allen-Zhu, Yuanzhi Li, Shean Wang, Lu~Wang, and Weizhu Chen.
\newblock Lora: Low-rank adaptation of large language models.
\newblock {\em arXiv preprint arXiv:2106.09685}, 2021.

\bibitem{brown2020language}
Tom Brown, Benjamin Mann, Nick Ryder, Melanie Subbiah, Jared~D Kaplan, Prafulla Dhariwal, Arvind Neelakantan, Pranav Shyam, Girish Sastry, Amanda Askell, et~al.
\newblock Language models are few-shot learners.
\newblock {\em Advances in neural information processing systems}, 33:1877--1901, 2020.

\bibitem{jiang2024mixtral}
Albert~Q Jiang, Alexandre Sablayrolles, Antoine Roux, Arthur Mensch, Blanche Savary, Chris Bamford, Devendra~Singh Chaplot, Diego de~las Casas, Emma~Bou Hanna, Florian Bressand, et~al.
\newblock Mixtral of experts.
\newblock {\em arXiv preprint arXiv:2401.04088}, 2024.

\bibitem{raposo2024mixture}
David Raposo, Sam Ritter, Blake Richards, Timothy Lillicrap, Peter~Conway Humphreys, and Adam Santoro.
\newblock Mixture-of-depths: Dynamically allocating compute in transformer-based language models.
\newblock {\em arXiv preprint arXiv:2404.02258}, 2024.

\bibitem{elhoushi2024layer}
Mostafa Elhoushi, Akshat Shrivastava, Diana Liskovich, Basil Hosmer, Bram Wasti, Liangzhen Lai, Anas Mahmoud, Bilge Acun, Saurabh Agarwal, Ahmed Roman, et~al.
\newblock Layer skip: Enabling early exit inference and self-speculative decoding.
\newblock {\em arXiv preprint arXiv:2404.16710}, 2024.

\bibitem{kosinski2023theory}
Michal Kosinski.
\newblock Theory of mind may have spontaneously emerged in large language models.

\bibitem{lecun2022path}
Yann LeCun.
\newblock A path towards autonomous machine intelligence version 0.9. 2, 2022-06-27.
\newblock {\em Open Review}, 62(1), 2022.

\bibitem{wei2022chain}
Jason Wei, Xuezhi Wang, Dale Schuurmans, Maarten Bosma, Fei Xia, Ed~Chi, Quoc~V Le, Denny Zhou, et~al.
\newblock Chain-of-thought prompting elicits reasoning in large language models.
\newblock {\em Advances in neural information processing systems}, 35:24824--24837, 2022.

\bibitem{goyal2023think}
Sachin Goyal, Ziwei Ji, Ankit~Singh Rawat, Aditya~Krishna Menon, Sanjiv Kumar, and Vaishnavh Nagarajan.
\newblock Think before you speak: Training language models with pause tokens.
\newblock {\em arXiv preprint arXiv:2310.02226}, 2023.

\bibitem{liu2023dropout}
Zhuang Liu, Zhiqiu Xu, Joseph Jin, Zhiqiang Shen, and Trevor Darrell.
\newblock Dropout reduces underfitting.
\newblock In {\em International Conference on Machine Learning}, pages 22233--22248. PMLR, 2023.

\bibitem{leech1992100}
Geoffrey~Neil Leech.
\newblock 100 million words of english: The british national corpus (bnc).

\bibitem{penedo2024refinedweb}
Guilherme Penedo, Quentin Malartic, Daniel Hesslow, Ruxandra Cojocaru, Hamza Alobeidli, Alessandro Cappelli, Baptiste Pannier, Ebtesam Almazrouei, and Julien Launay.
\newblock The refinedweb dataset for falcon llm: Outperforming curated corpora with web data only.
\newblock {\em Advances in Neural Information Processing Systems}, 36, 2024.

\bibitem{sennrich2015neural}
Rico Sennrich, Barry Haddow, and Alexandra Birch.
\newblock Neural machine translation of rare words with subword units.
\newblock {\em arXiv preprint arXiv:1508.07909}, 2015.

\bibitem{schuster2012japanese}
Mike Schuster and Kaisuke Nakajima.
\newblock Japanese and korean voice search.
\newblock {\em 2012 IEEE International Conference on Acoustics, Speech and Signal Processing (ICASSP)}, pages 5149--5152, 2012.

\end{thebibliography}

\appendix
\section{Overview of LLM Architectures}
\label{sec:llm_related}
Large language models (LLMs) have become a cornerstone of modern natural language processing (NLP) due to their ability to understand and generate human-like text. At the heart of these models lies the transformer architecture, introduced by \cite{vaswani2017attention}, which has since become the standard for NLP tasks. The transformer architecture is composed of several key components:
\begin{figure}
    \centering
    \includegraphics[width=0.8\linewidth]{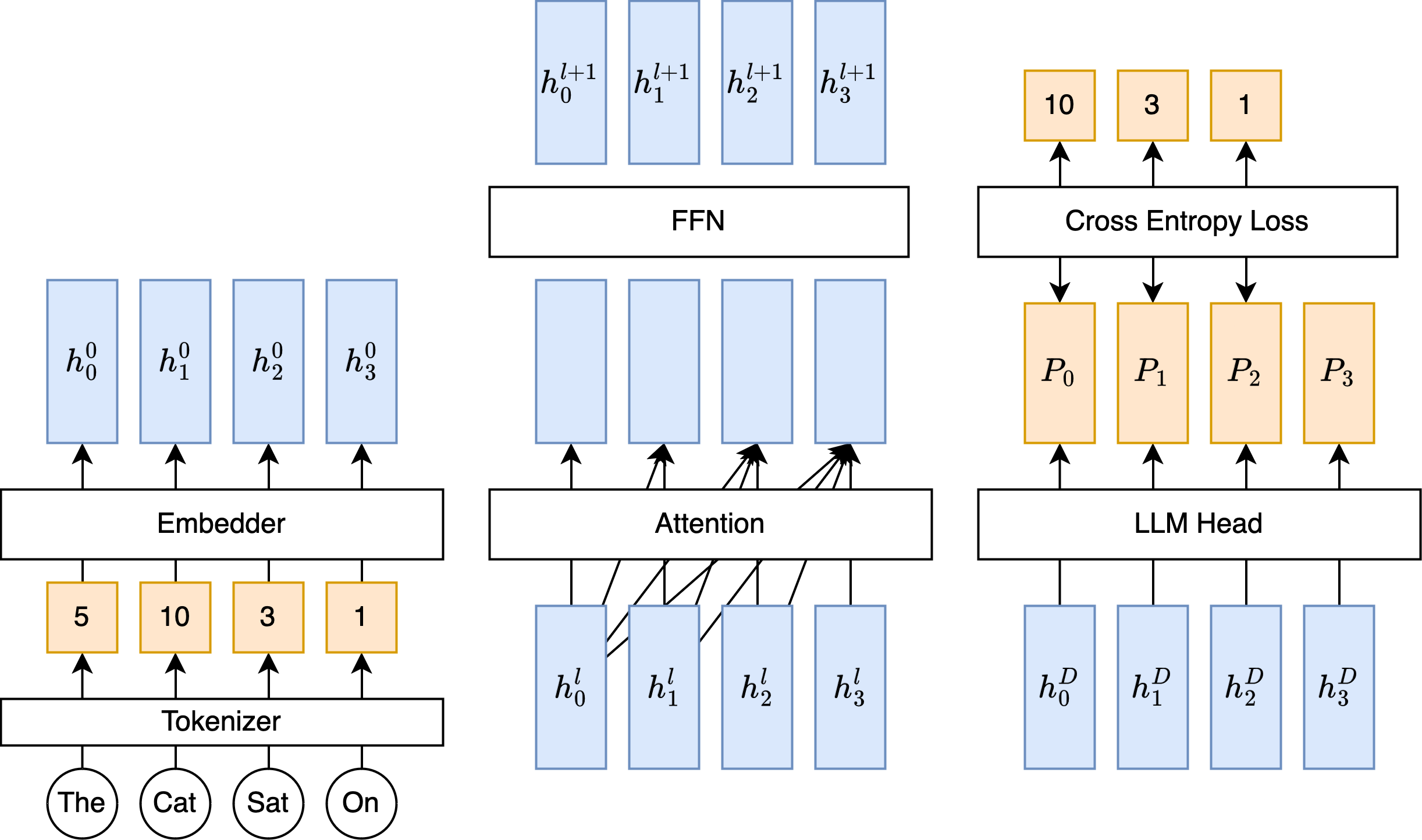}
    \caption{Diagram demonstrating flow of information through transformer components during training}
    \label{fig:tansformer-overview}
\end{figure}
\subsection{Tokenization}
The process begins with tokenization, where input text is split into smaller units called tokens. These tokens can be words, subwords, or even characters, depending on the chosen tokenizer. Common tokenization methods include Byte Pair Encoding (BPE) \cite{sennrich2015neural} and WordPiece \cite{schuster2012japanese}, which help in handling the vast vocabulary of natural languages by breaking down rare words into more frequent subwords.

\subsection{Embeddings}
\paragraph{Embedding Lookup Tables} After tokenization, the token indices are typically use to index an embedding matrix $W_e$ of shape $N\times E$ where $N$ is the vocab size and $E$ is the embedding dimension.
\paragraph{Positional Embeddings} Unless explicitly given, by default a transformer would have no way of distinguishing adjacent tokens from distant tokens. As such this information has to be encoded into the model's embeddings. In the original transformer paper\cite{vaswani2017attention}, this was done in terms of Sine/Cosine embeddings, but other methods include learnt positional embeddings and more recently RoPE Embeddings\cite{su2024roformer} which modify the hidden states directly in the attention layers.
\subsection{Transformer Blocks}
The core of an LLM consists of multiple transformer blocks, each containing two main sub-layers: multi-head self-attention mechanisms and feed-forward neural networks (FFNs).

\paragraph{Self-Attention Mechanism} This mechanism allows the model to weigh the importance of different tokens in the input sequence when making predictions. By considering the entire sequence, the model can capture long-range dependencies and contextual information effectively. The self-attention mechanism is computed as follows:
\begin{equation*}
    \text{Attention}(Q, K, V) = \text{softmax}\left(\frac{QK^T}{\sqrt{d_k}}\right)V
\end{equation*}
where $Q$, $K$, and $V$ are the query, key, and value matrices repsectively and $d_k$ is the dimension of the key vectors.

\paragraph{Feed-Forward Neural Networks (FFNs)} Following the self-attention mechanism, each token's representation is passed through a feed-forward neural network. At least in the original transformer design, this sub-layer consists of two linear transformations with a non-linear activation function in between, further refining the token representations:
\begin{equation*}
    \text{FFN}(x) = \sigma(xW_1 + b_1)W_2 + b_2
\end{equation*}
where $W_1$ and $W_2$ are weight matrices, $b_1$ and $b_2$ are bias vectors, and $\sigma$ is a non-linear activation function. More recent transformers like \cite{touvron2023llama} forgo the bias and use the SwiGLU feed forward network\cite{shazeer2020glu} given by:
\begin{equation*}
    \text{SwiGLU}(x) = (\text{Swish}(xW) \otimes xV)W_2
\end{equation*}
Where $\otimes$ is the hadamard or elementwise product, $W,V$ are are additional weight matrices, and $\text{Swish}(x)=\sigma(x)\cdot x$ where $\sigma$ is the sigmoid function.

\paragraph{Next Token Prediction Head} At the final stage of the transformer model, the refined token representations are used to predict the next token in the sequence. This is achieved by applying a linear transformation followed by a softmax function, which generates a probability distribution over the vocabulary, indicating the likelihood of each possible next token:
\begin{equation*}
    P(x_i | x_{<i}) = \text{softmax}(h_i W_e^T)
\end{equation*}
where $h_i$ is the hidden state for the $i$-th token and $W_e$ is the embedding matrix. The model is typically trained using the cross-entropy loss function, defined as:
\begin{equation*}
    \mathcal{L} = -\sum_{i} \log P(x_i | x_{<i})
\end{equation*}

These components work together to enable LLMs to perform a wide range of NLP tasks with high accuracy and fluency. However, the immense size of these models, often containing billions of parameters, poses significant challenges in terms of computational requirements and energy consumption.

\end{document}